\documentclass[conference]{IEEEtran}
\IEEEoverridecommandlockouts

\usepackage{cite}
\usepackage{amsmath,amssymb,amsfonts}
\usepackage{algorithmic}
\usepackage{graphicx}
\usepackage{textcomp}
\usepackage{hyperref}
\usepackage{pifont}
\usepackage{tabularx}
\usepackage{tcolorbox}
\usepackage{booktabs}
\newtheorem{definition}{Definition}
\usepackage{xcolor}
\def\BibTeX{{\rm B\kern-.05em{\sc i\kern-.025em b}\kern-.08em
    T\kern-.1667em\lower.7ex\hbox{E}\kern-.125emX}}

\def\BibTeX{{\rm B\kern-.05em{\sc i\kern-.025em b}\kern-.08em
    T\kern-.1667em\lower.7ex\hbox{E}\kern-.125emX}}
\definecolor{posgray}{gray}{0.95} 
\begin{document}

\title{Vision Paper:
Why Distributed LLM Post-Training and Decentralized Alignment Are Now Essential
}

\author{\IEEEauthorblockN{Zhanhong Jiang, Qisai Liu, Soumik Sarkar}
\IEEEauthorblockA{
\textit{Iowa State University,}
Ames, USA \\
\{zhjiang, supersai, soumiks\}@iastate.edu}
}

\maketitle

\begin{abstract}
This vision paper argues that distributed learning, including federated and decentralized, should become a first‑class paradigm in LLM post‑training, not merely a privacy add‑on. Current methods like RLHF and DPO rest on an invisible assumption: preference data should be centralized. We contend this assumption is scientifically unjustified, practically unsustainable, and ethically problematic. As LLMs deploy into healthcare, law, finance, and multilingual communities, preference data originates from populations that cannot or will not centralize. We develop a three‑pillar argument grounded in (1) privacy and data sovereignty, (2) pluralism and representational equity, and (3) robustness and scalability. We formalize key technical challenges unique to this intersection, reference drift, preference heterogeneity, and the DP‑utility trade‑off for contrastive losses, and propose concrete algorithmic directions, and sketch a research agenda. Our analysis shows distributed post‑training is not only necessary but tractable, and ignoring it risks AI that is legally noncompliant, representationally narrow, and brittle at scale.
\end{abstract}


\section{Introduction}
\label{sec:intro}

The alignment of large language models (LLMs) with human preferences has become a cornerstone of modern machine learning. Reinforcement learning from human feedback (RLHF)~\cite{ ouyang2022training} and its more efficient successor, direct preference optimization (DPO)~\cite{rafailov2023direct}, have enabled the transformation of raw pretrained models into systems that are helpful, harmless, and honest. A growing family of methods, KTO~\cite{ethayarajh2024kto}, ORPO~\cite{hong2024reference}, and SimPO~\cite{meng2024simpo}, has further refined the preference optimization landscape. These advances represent genuine scientific progress and have powered chatbots, coding assistants, and creative tools~\cite{afzoon2025modeling} used by hundreds of millions of people. 
Yet virtually all of this progress rests on an assumption that has gone largely unquestioned:

\begin{tcolorbox}[colback=posgray, boxrule=0.8pt, arc=3pt, left=6pt, right=6pt, top=4pt, bottom=4pt]
\textbf{The Centralization Assumption.} \textit{Alignment requires that preference data, i.e., human judgments about model outputs, be gathered, stored, and processed in a central location under the control of the model developer. Reward models are trained centrally, policy gradients are aggregated centrally, and DPO losses are minimized centrally.}
\end{tcolorbox}

\vspace{4pt}
This centralization assumption is so deeply embedded in current practice that it has become invisible. But it is rapidly colliding with three hard realities of real‑world LLM deployment. First, privacy and data sovereignty: preference data often encodes intimate values, medical contexts, or legal deliberations, and regulations such as the EU AI Act~\cite{edwards2021eu} increasingly restrict or prohibit the transfer of such data to a central server. Second, pluralism and representational equity: centralized annotation pipelines systematically overrepresent English‑language, Western, and high‑income perspectives~\cite{santy2023nlpositionality}, producing models that perform unevenly across global populations. Third, robustness and scalability: centralized pipelines face super‑linear growth in annotation costs and present concentrated attack surfaces for preference poisoning~\cite{wan2023poisoning}. These pressures are not hypothetical. They are already shaping litigation, product development, and regulatory enforcement. We argue that the centralization assumption is no longer scientifically, ethically, or practically sustainable, and that the ML community should embrace a fundamentally different paradigm.

\begin{figure}
    \centering
    \includegraphics[width=\linewidth]{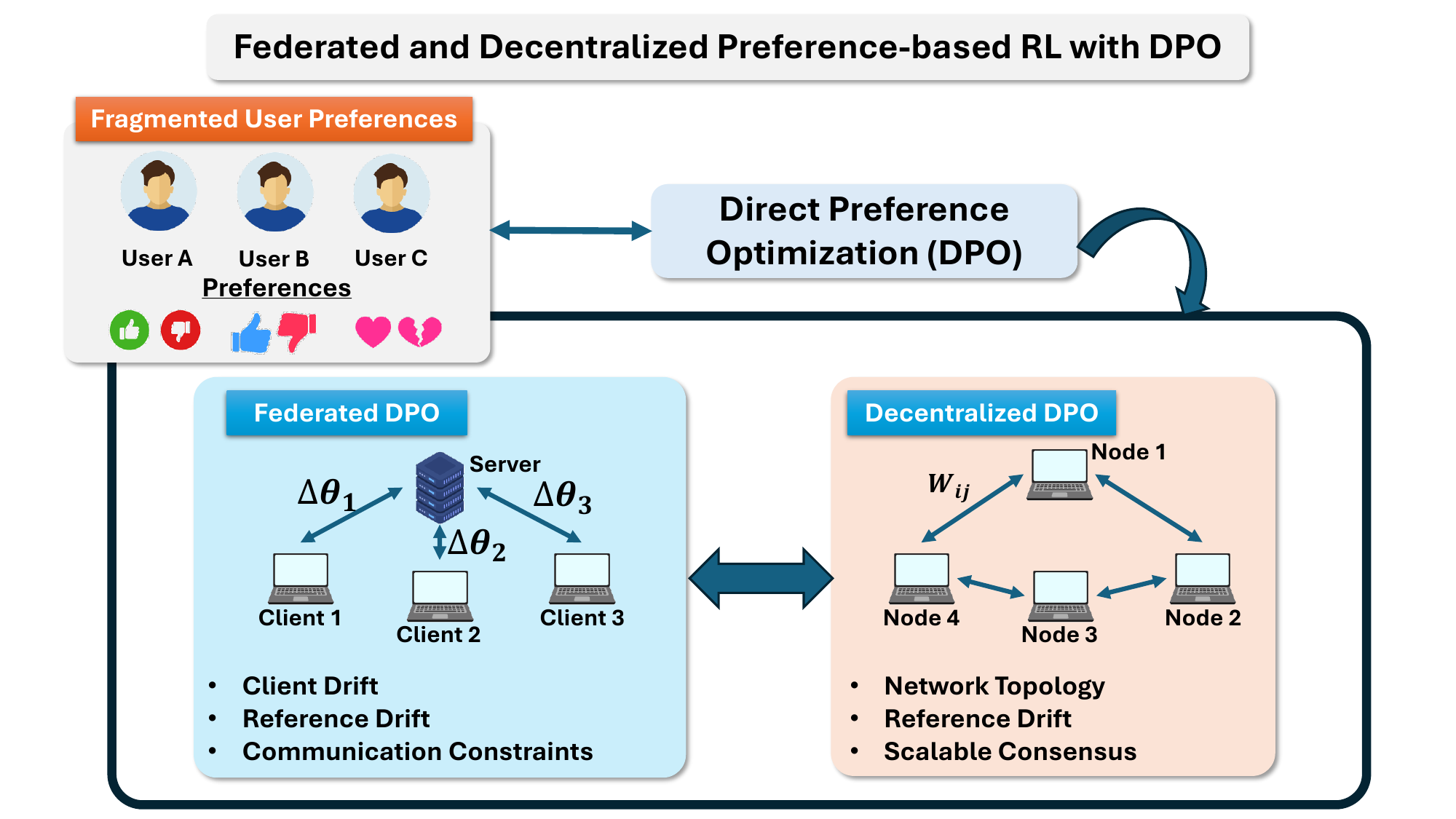}
    \caption{Two distributed post-training paradigms. \textbf{(a) Federated DPO}: clients send local model updates (optionally privatized via differential privacy) $\Delta\theta_i$ to a trusted aggregation server, which broadcasts the averaged policy $\bar\theta$. \textbf{(b) Decentralized DPO}: no server exists; nodes exchange models with neighbors via mixing weights $W_{ij}$ and update locally. Reference drift (see Section~\ref{sec:technical}) affects both paradigms and must be mitigated by synchronized reference resets.}
    \label{fig:paradigms}
\end{figure}

\textbf{The Vision.}
\textbf{We argue that the centralization assumption is scientifically
unjustified, practically unsustainable, and ethically
problematic, and that combining distributed learning (federated and
decentralized as in Figure~\ref{fig:paradigms}) with LLM post-training algorithms such as DPO is not
merely feasible but \emph{crucial} to the long-term scientific
integrity and societal legitimacy of AI alignment.} 

The argument rests on three reinforcing pillars, each of which is developed in detail in Section~\ref{sec:pillars}.
We do not claim that distributed alignment is easier than centralized alignment. To the contrary, it introduces novel technical challenges, including reference drift, preference heterogeneity, DP‑utility trade‑offs for contrastive losses, and Byzantine robustness in preference space, that we formalize in Section~\ref{sec:technical}. Our central claim is that these challenges are tractable and that the ML community has so far neglected them due to an unexamined commitment to centralization.

\textbf{Contributions.}
This paper makes four interconnected contributions. First, we build a three‑pillar (Section~\ref{sec:pillars}) normative argument showing that the centralization assumption is simultaneously a legal liability (under GDPR, HIPAA, and the EU AI Act), an equity failure (systematic underrepresentation of non‑Western annotators), and a scalability ceiling, with pillar being independently falsifiable. Second, we formally derive federated DPO (FedDPO) and introduce a novel decentralized DPO (D‑DPO) that replaces server aggregation with gossip‑based mixing (Section~\ref{sec:technical}). In doing so, we identify \emph{reference drift} as a critical failure mode unique to distributed preference optimization and propose the Synchronized Reference Update (SRU) protocol to mitigate it. Third, we provide a structured taxonomy of five open challenges, including reference drift, preference heterogeneity, communication efficiency, DP‑utility trade‑offs, and Byzantine robustness, distinguishing which have partial solutions and which remain genuinely open, along with concrete algorithmic directions for each. Finally, we bridge three previously disconnected literatures (federated LLM fine‑tuning, pluralistic alignment, and privacy‑preserving ML) to reveal a gap: no existing work treats distributed preference optimization as a unified problem. We then outline seven prioritized research directions from convergence theory to governance to close that gap (Section~\ref{sec:agenda}).

\section{Related Work}
\label{sec:related}

\textbf{Federated fine-tuning of LLMs.} Recent surveys~\cite{wu2025survey} show growing interest in applying federated learning~\cite{mcmahan2017communication} to LLM fine-tuning. FedIT~\cite{zhang2023towards} and FedGPT~\cite{yao2024federated} demonstrate federated instruction tuning at modest scale (up to several billion parameters). Parameter-efficient fine-tuning (PEFT) methods, notably LoRA~\cite{hu2022lora} and QLoRA~\cite{dettmers2023qlora}, reduces communication overhead by restricting updates to low‑rank adapters, decreasing trainable parameters by over 99\% and per‑round transmission to hundreds of MB for 7B‑scale models. A few recent works have begun exploring federated preference optimization. For example, comparing KTO and DPO in federated settings~\cite{spadea2025federated}, proposing personalized DPO for client‑specific preferences~\cite{zhu2026fedpdpo}, and using DPO to generate differentially private synthetic data~\cite{hou2025popri}, but these efforts do not address the full set of challenges we tackle, such as reference drift, DP‑contrastive trade‑offs, and fully decentralized alignment.
Critically, \emph{the vast majority of existing federated LLM fine-tuning work targets supervised objectives (cross-entropy on instruction-response pairs)}. Federated preference optimization with its contrastive loss structure and explicit reference policy dependency remains largely absent from the literature. This gap is the primary focus of our work.

\textbf{Pluralistic and personalized alignment.} S{\o}rensen et al.~\cite{sorensen2024roadmap} taxonomize pluralistic alignment into three families: overton models that represent the distribution of human values, steerable models conditioned on preference embeddings, and personalized models that adapt to individual users. Concurrently, personalized FL methods~\cite{fallah2020personalized} decompose model parameters into shared and client-specific components to handle data heterogeneity. The intersection of these two lines of work is underexplored: existing personalized FL theory does not account for contrastive preference losses (e.g., DPO’s dependence on pairwise comparisons), and pluralistic alignment research does not consider the privacy, communication, or data‑locality constraints of distributed settings. Bridging this gap would enable federated pluralistic alignment, where a global model accommodates diverse value systems without centralizing sensitive preference data.

\textbf{Privacy-preserving alignment.}
Differential privacy (DP) for LLM fine‑tuning has been studied primarily in the supervised setting~\cite{li2022large}, where large models exhibit surprising robustness to DP noise due to over‑parameterization and gradient redundancy. Whether this robustness transfers to preference optimization is unknown. The DPO gradient (Eq.~\ref{gradient_dpo})is a difference of two log‑probability gradients weighted by a sigmoid; this structure has lower intrinsic signal‑to‑noise ratio than a standard cross‑entropy gradient, especially when the policy does not yet separate preferred from dispreferred responses. Consequently, DP noise may disproportionately degrade alignment quality. Secure aggregation~\cite{bonawitz2017practical} and multi-party computation offer complementary cryptographic protections but have not been integrated with preference optimization pipelines. Doing so would require new protocols that operate on contrastive losses.

\textbf{Robustness to alignment poisoning.} Wan et al.~\cite{wan2023poisoning} show that poisoning as little as 0.5\% of instruction‑tuning data can significantly alter model behavior. Byzantine‑robust aggregation methods, e.g., Krum~\cite{blanchard2017machine} and coordinate‑wise median~\cite{yin2018byzantine}, provide partial defenses but rely on the assumption that honest clients’ gradients are similar. This assumption fails when clients have legitimately heterogeneous preferences (e.g., different cultural norms or domain‑specific values). In such cases, a malicious client’s update may be statistically indistinguishable from that of an honest client with an unusual but genuine preference distribution. Distinguishing adversarial poisoning from benign preference divergence is an open problem with no current solution. We return to it in Section~\ref{subsec:byzantine}.

\textbf{Positioning relative to prior work.} Our contribution is distinct from all of the above. We provide the first unified argument that distributed post‑training is \emph{necessary} (not merely useful) on legal, representational, and scalability grounds. We formally characterize problems unique to the distributed preference optimization setting with reference drift (Section~\ref{subsec:refdrift}) and the DP‑contrastive trade‑off (Section~\ref{dp_contrastive}), and propose concrete algorithmic mitigations. Finally, we outline a structured research agenda that connects federated learning, pluralistic alignment, privacy, and robustness into a coherent framework for distributed LLM alignment. 

\section{Background}
\label{sec:background}

\textbf{LLM Post-Training and Preference Optimization.}
Modern LLM deployment follows a two-phase paradigm: pretraining on large corpora, then post-training to align with human intent. RLHF~\cite{ouyang2022training} frames alignment as a KL-constrained reward maximization problem. Given a reward model $r_\phi$ parameterized by $\phi\in\mathbb R^D$ and trained on preference pairs $(x, y_w, y_l)$ ($x$ is the input prompt, $y_w$ is the preferred response, and $y_l$ is the dispreferred response) via the Bradley-Terry model~\cite{bradleyterry1952}, the policy is optimized as:
$
\max_{\pi_\theta} \mathbb{E}_{x,\, y \sim \pi_\theta}\!\left[r_\phi(x, y) - \beta \log \frac{\pi_\theta(y|x)}{\pi_\text{ref}(y|x)}\right],
$
where $\pi_\theta$ is the policy model parameterized by $\theta\in\mathbb R^d$, and $\pi_\text{ref}$ is the reference policy model. $\beta$ is the inverse temperature parameter.
The closed-form optimal solution~\cite{rafailov2023direct} is $\pi^*(y|x) \propto \pi_\text{ref}(y|x)\exp\!\left(r(x,y)/\beta\right)$, implying the implicit reward $r^*(x,y) = \beta\log(\pi^*(y|x)/\pi_\text{ref}(y|x)) + \beta\log Z(x)$, where $Z(x)=\sum_y\pi_\text{ref}(y|x)\text{exp}(r(x,y)/\beta)$. DPO substitutes this reparameterization directly into the preference loss, yielding a reward-model-free loss:
$
\mathcal{L}_\text{DPO}(\pi_\theta;\,\pi_\text{ref}) = -\mathbb{E}_{(x, y_w, y_l)\sim \mathcal{D}}\bigg[\log \sigma\bigg(\beta \log \frac{\pi_\theta(y_w|x)}{\pi_\text{ref}(y_w|x)} - \beta \log \frac{\pi_\theta(y_l|x)}{\pi_\text{ref}(y_l|x)}\bigg)\bigg].
$
All current preference optimization methods, e.g., DPO, IPO, KTO, SimPO, assume that $\mathcal{D}$ and $\pi_\text{ref}$ are globally accessible. This is the assumption we challenge.

\textbf{Distributed Learning: Federated and Decentralized.}
We first introduce the federated learning (FL) setting. In the canonical cross-silo
setup~\cite{mcmahan2017communication,kairouz2021advances}, $N$
clients hold disjoint local datasets $\{\mathcal{D}_i\}_{i=1}^N$.
A central server coordinates training via FedAvg:
\begin{equation}
  \theta^{(t+1)} = \sum_{i=1}^N w_i\,\theta_i^{(t)},\quad
  w_i = \frac{|\mathcal{D}_i|}{\sum_j|\mathcal{D}_j|},
  \label{eq:fedavg}
\end{equation}
where $\theta_i^{(t)}$ results from $K$ local gradient steps.
FL preserves raw data locality while enabling model-level
collaboration.
Decentralized
SGD~\cite{lian2017can,koloskova2020unified, saadati2024dimat,saadati2025decaf} removes the server
entirely.  Clients communicate over a peer-to-peer graph
$G=(V,E)$ with doubly-stochastic mixing matrix
$W\in\mathbb{R}^{N\times N}$, where $W_{ij}>0$ iff $(i,j)\in E$.
The update rule is:
\begin{equation}
  \theta_i^{(t+1)}
    = \sum_{j\in\mathcal{N}(i)} W_{ij}\,\theta_j^{(t)}
      -\eta\,\nabla_\theta\mathcal{L}^i_\text{DPO}(\theta_i^{(t)}),
  \label{eq:decsgd}
\end{equation}
where $\mathcal{N}(i)$ denotes the neighbors of node $i$ in $G$, which is defined by $\mathcal{N}(i)=\{j|(i,j)\in E\;\text{or}\;i=j\}$. $\nabla_\theta\mathcal{L}^i_\text{DPO}(\theta_i^{(t)})$ is the local gradient defined in Eq.~\ref{gradient_dpo}.
Convergence rates depend on the spectral gap $1-\lambda_2(W)>0$
of the mixing matrix.  Both paradigms face \emph{statistical
heterogeneity} (non-IID local distributions), \emph{systems
heterogeneity} (varying compute and bandwidth), and server
trust constraints in FL---challenges well-characterised for
supervised objectives~\cite{kairouz2021advances} but whose
interaction with contrastive preference losses introduces
fundamentally new difficulties explored in
Section~\ref{sec:technical}.

\section{The Three-Pillar Argument}
\label{sec:pillars}

\textbf{Pillar I: Privacy and Data Sovereignty.}
Human preference data is not merely sensitive---it is intimate and legally protected. When a clinician rates a model's response to a diagnostic query, the rating may be entangled with protected health information (PHI). When a legal professional evaluates a model's statutory interpretation, the judgment may reflect privileged deliberation. When a citizen evaluates a model's political commentary, the response encodes personal values that fall within the scope of opinion privacy.
The regulatory environment has evolved rapidly. GDPR's data minimization principle restricts collecting preference data beyond what is necessary for a specified purpose; GDPR's purpose limitation principle restricts using data collected for one purpose (e.g., product improvement) for another (alignment training). HIPAA governs any preference data collected in healthcare contexts. The EU AI Act classifies certain LLM deployments as high-risk systems subject to additional data governance requirements. Brazil's LGPD~\cite{canedo2021proposal} and China's PIPL~\cite{tan2021china} impose analogous national frameworks. Emerging AI-specific regulation in the US, UK, and Canada will tighten these constraints further.

Centralized RLHF and DPO may, in many deployment contexts, create significant legal risk or be difficult to reconcile with existing regulations. Actual compliance depends on implementation details (e.g., de‑identification, consent mechanisms, data processing agreements) and jurisdictional interpretation. Nevertheless, the architectural tension is clear: shipping raw preference data to a central server invites regulatory scrutiny under GDPR, HIPAA, and similar frameworks.
The act of shipping preference data to a central server for training may constitute unauthorized data transfer, purpose limitation violation, or cross-border data export depending on jurisdiction. This is not a future concern---it is a present constraint that model developers are navigating through data use agreements of uncertain enforceability.
Federated and decentralized learning, augmented by differential privacy~\cite{dwork2014algorithmic} and secure aggregation~\cite{bonawitz2017practical}, offers a principled remedy: preference data never leaves the client, only privatized model updates do. This is not only more convenient, but also architecturally compliant.

\textbf{Pillar II: Pluralism and Representational Equity.}
The outputs of centralized alignment reflect the inputs of centralized annotation. Documented analyses~\cite{bender2021dangers, santy2023nlpositionality} reveal that English-language, Western, high-income perspectives are systematically overrepresented in preference datasets underlying frontier models. The practical consequence is models that perform well for a narrow demographic while producing lower-quality, culturally inappropriate, or actively harmful outputs for others.
The root cause is structural and not easily fixed within a centralized paradigm. Large-scale preference collection requires annotator pools that can be recruited, quality-controlled, and compensated centrally, which is a process that gravitates toward geographic and demographic concentration. Crowdsourcing platforms compound this by drawing workers from specific economic contexts. Even well-resourced multilingual annotation efforts face the practical impossibility of representing the full breadth of human linguistic and cultural diversity at the scale required for frontier alignment.
Distributed alignment essentially changes the structural incentives. If post-training can be performed in a distributed manner, any institution or community with locally generated preference data can participate in alignment \emph{without surrendering data control}. A rural hospital in Kenya, a legal clinic in Bangladesh, a community media organization in Bolivia. Each could contribute domain-specific preference signal to a globally shared alignment process while retaining full data sovereignty. The global alignment would be better calibrated precisely because the distributional coverage is broader. This is not a diversity benefit layered on top of a sound system. It is a prerequisite for a system that is genuinely globally useful.

\textbf{Pillar III: Robustness and Scalability.}
Centralized post-training faces a scalability ceiling that is already visible. As the scope of alignment expands, e.g., more models, more domains, more deployment contexts, more languages, more users, the volume of preference data, the number of annotators, and the organizational complexity of centralized pipelines grow super-linearly. The annotation infrastructure required to align a single frontier model already involves thousands of contractors; scaling to the diversity of use cases envisioned for future AI systems is implausible within a centralized paradigm.
Distributed architectures are inherently more scalable: adding clients increases both the pool of preference data and the computational resources available for local training, without requiring proportional expansion of central infrastructure.
Centralized systems also present concentrated attack surfaces. A single poisoning attack on a central preference dataset~\cite{wan2023poisoning} or reward model can corrupt the alignment of all downstream deployments. The 2023 Anthropic/OpenAI API incidents~\cite{sidorkin2025ai} and multiple documented cases of RLHF reward hacking illustrate the brittleness of centralized pipelines when reward model assumptions are violated. In distributed systems, adversarial clients can influence only their local contribution, and robust aggregation methods~\cite{yin2018byzantine} can bound their impact. While the failure mode of distributed systems is graceful degradation, the failure mode of centralized systems is catastrophic collapse.
Together, these arguments make a strong normative case for distributed post‑training. However, a normative case alone is insufficient, and the vision is credible only if the technical obstacles are precisely understood and tractable paths forward exist. The next section formalizes the core challenges unique to distributed preference optimization, including reference drift, preference heterogeneity, communication efficiency, DP‑contrastive trade‑offs, and Byzantine robustness, and proposes concrete algorithmic mitigations.

\section{Technical Landscape: Formalizing Distributed Post-Training}
\label{sec:technical}

We formalize both the federated and decentralized DPO problems and
characterize their principal technical challenges.  This is the
technical core of our argument: the vision is credible only if the
obstacles are precisely understood and tractable paths forward can be
identified.
 
\textbf{Federated DPO (FedDPO).} 
Client $i$ holds local preference data
$\mathcal{D}_i=\{(x^{(k)}_i,y^{(k)}_{w,i},y^{(k)}_{l,i})\}_{k=1}^{n_i}$
drawn from client-specific distribution $\mathcal{P}_i$.  The global
FedDPO objective is:
\begin{equation}
  \min_\theta\;\mathcal{L}_\text{FedDPO}(\theta)
  = \sum_{i=1}^N w_i\,\mathcal{L}_{\text{DPO}}^i(\pi_\theta;\pi_\text{ref}),
  \quad w_i = n_i\big/\!\sum_j n_j.
  \label{eq:feddpo}
\end{equation}
Each client performs $K$ local gradient steps on
$\mathcal{L}_{\text{DPO}}^i$ before uploading the LoRA adapter
increment $\Delta\theta_i$ to the server, which computes the
weighted average in Eq.~\eqref{eq:fedavg}.  The local DPO gradient is:
\begin{equation}\label{gradient_dpo}
\begin{split}
  &\nabla_\theta\mathcal{L}_{\text{DPO}}^i
  = -\beta\,\mathbb{E}_{(x,y_w,y_l)\sim\mathcal{D}_i}\bigg[
    \sigma\!\left(\hat{h}_{\theta}^i\right)
    \bigl(\nabla_\theta\log\pi_\theta(y_w|x)
         \\&-\nabla_\theta\log\pi_\theta(y_l|x)\bigr)
  \bigg],
\end{split}
\end{equation}
where
$\hat{h}_{\theta}^i=\beta\log\!\tfrac{\pi_\theta(y_w|x)}{\pi_\text{ref}(y_w|x)}
-\beta\log\!\tfrac{\pi_\theta(y_l|x)}{\pi_\text{ref}(y_l|x)}$.
The gradient is a reweighted difference of log-probability
gradients: the sigmoid weight $\sigma(\hat{h})$ is large (gradient
is informative) when the current policy does \emph{not} yet
distinguish preferred from dispreferred responses, and collapses to
zero once the policy separates them perfectly.  This structure
introduces challenges absent from standard supervised federated
fine-tuning, such as the reference drift and increased sensitivity to client heterogeneity.
 
\textbf{Decentralized DPO (D-DPO).}
In the fully decentralized setting there is no aggregation server.
Each node $i$ maintains its own local policy $\pi_{\theta_i}$ and
communicates only with graph neighbors.  The D-DPO update combines
gossip mixing with a local DPO gradient step:
\begin{equation}
  \theta_i^{(t+1)}
  = \sum_{j\in\mathcal{N}(i)} W_{ij}\,\theta_j^{(t)}
    - \eta\,\nabla_\theta\mathcal{L}_{\text{DPO}}^i
      \!\left(\pi_{\theta_i^{(t)}};\,\pi_{\text{ref},i}^{(t)}\right),
  \label{eq:ddpo}
\end{equation}
where $\pi_{\text{ref},i}^{(t)}$ is node $i$'s local reference
policy at round $t$, $\eta$ is the step size.  In the absence of a broadcast mechanism,
reference policies cannot be globally synchronized, making
reference drift (Section~\ref{subsec:refdrift}) structurally more
severe than in FedDPO.  When the mixing matrix $W$ is doubly
stochastic and the graph is connected, D-DPO converges to a
neighbourhood of the global minimum at a rate governed by
$O(1/\sqrt{T})$ for nonconvex losses~\cite{koloskova2020unified},
\emph{provided} the local losses are consistent, which is a condition
violated by reference drift.
A practical mitigation unique to the decentralized setting is
\emph{neighbourhood reference consensus}: node $i$ periodically
resets $\pi_{\text{ref},i}$ to the gossip average of its
neighbours' current policies, $\pi_{\text{ref},i}\leftarrow
\sum_{j\in\mathcal{N}(i)}W_{ij}\,\pi_{\theta_j}$.  
This respects the serverless constraint and bounds reference divergence, though formal analysis of consensus‑and‑align remains open. Reference‑free methods (SimPO, ORPO) eliminate the reference model entirely, sidestepping drift without synchronization overhead. However, they introduce new sensitivities: SimPO depends on prompt length distributions across clients, and ORPO may amplify local overfitting. Adapting them to decentralized settings is a promising direction (see R2, Section~\ref{sec:agenda}).

\subsection{Challenge 1: Reference Drift}\label{subsec:refdrift}

\begin{definition}[Reference Drift]\label{reference_drift}
Let client $i$ use reference model $\pi^{(i,t)}_\text{ref}$ at round $t$ while the server (or other clients) use $\pi^{(j,t)}_\text{ref}$. Reference drift is the divergence between the implicit reward functions induced by these references:
$
    \Delta r^{(t)}_{i,j}(x,y)=\beta\text{log}\frac{\pi_\theta(y|x)}{\pi^{(i,t)}_\text{ref}(y|x)}-\beta\text{log}\frac{\pi_\theta(y|x)}{\pi^{(j,t)}_\text{ref}(y|x)}.
$
When $\pi^{(i,t)}_\text{ref}\neq \pi^{(j,t)}_\text{ref}$, clients optimize different implicit rewards even when their policies $\pi_\theta$ are identical. A scalar summary is $\delta_i^{(t)}=\mathrm{KL}(\pi_{\theta_i}^{(t)}\|\pi_\text{ref}^{(t)})
-\mathrm{KL}(\pi_{\theta_i}^{(t)}\|\pi_\text{ref}^{(0)})$, measuring how far the effective KL penalty has shifted from the intended baseline.
\end{definition}
 
In standard federated fine-tuning the loss $\mathcal{L}_i(\theta)$
depends only on $\theta$. However, in FedDPO and D-DPO the loss
$\mathcal{L}_{\text{DPO}}^i(\pi_\theta;\pi_\text{ref})$ depends on
\emph{both} the policy and the reference.  If clients maintain
different effective references, they are optimizing different
implicit rewards, and gradient aggregation conflates incommensurable
objectives.
 
 

\textbf{Illustrative counterexample.} Consider two clients with opposite preferences on a single prompt–response pair $(x,y_w,y_l)$. Client 1 has reference $\pi^1_\text{ref}$ that already slightly favors $y_w$; client 2 has reference $\pi^2_\text{ref}$ that already slightly favors $y_l$. Both use the same policy $\pi_\theta$ initialized near uniform. Their local DPO gradients point in nearly opposite directions. Under FedAvg, the aggregated gradient is near zero, causing the global policy to stall even though each client could make progress individually. This stagnation does not occur when $\pi^1_\text{ref}=\pi^2_\text{ref}$. The example demonstrates that reference drift is not merely a theoretical nuisance but can practically halt convergence.

\textbf{Mitigation (FedDPO---SRU).}  We propose \emph{Synchronized
Reference Update} (SRU): every $T_\text{ref}$ global rounds, the
server broadcasts the current global policy $\bar\theta$ as the new
reference, and all clients reset accordingly.  This bounds
$\delta_i^{(t)}\leq O(K\eta)$ between resets.  An adaptive variant
sets $T_\text{ref}$ by monitoring the variance
$\mathrm{Var}_i(\hat{h}^i_{\theta})$ across clients: large variance
signals diverging references.
 
\textbf{Mitigation (D-DPO---Neighbourhood Consensus).}  Without a
server, each node resets its reference to the gossip average of
neighbours' policies every $T_\text{ref}$ local steps, as described
in Section~\ref{sec:technical}.  Both mitigations are compatible with
LoRA-only communication: only the adapter increments, not the full
model, need to be exchanged.

\subsection{Challenge 2: Preference Heterogeneity}\label{subsec:heterogeneity}

Client datasets $\{\mathcal{D}_i\}$ reflect heterogeneous, potentially conflicting preference distributions $\{\mathcal{P}_i\}$. The gradient of the federated objective~\eqref{eq:feddpo} is $\sum_i w_i \nabla_\theta \mathcal{L}_{\text{DPO}}^i$. When $\mathcal{P}_i$ and $\mathcal{P}_j$ place conflicting preference orderings on the same $(x, y_w, y_l)$ triple, their local gradients may be anti-parallel, and FedAvg produces a global model that satisfies no client well, the ``client drift'' phenomenon familiar from heterogeneous FL~\cite{kairouz2021advances}, exacerbated here by the contrastive loss structure.

\subsection{Challenge 3: Communication Efficiency}

Distributing DPO over a 70B model at FP32 would require transmitting
$\sim$280\,GB per round per client---clearly infeasible.
LoRA~\cite{hu2022lora} resolves this by restricting updates to
low-rank matrices $\Delta W=BA$ with $r\ll\min(d_\text{in},d_\text{out})$,
reducing per-layer communication from $O(d_\text{in}d_\text{out})$
to $O(r(d_\text{in}+d_\text{out}))$.  At $r=16$ over a 7B model
this achieves a ${\sim}2000{\times}$ reduction, making distributed
rounds comparable in bandwidth to distributed training of
discriminative models.  QLoRA~\cite{dettmers2023qlora} extends
this with 4-bit quantization, further halving memory footprint on
client devices.
An important open question is how restriction to a low-rank
subspace interacts with the contrastive DPO loss: DPO requires
the policy to simultaneously increase the log-probability of $y_w$
\emph{and} decrease that of $y_l$, potentially requiring orthogonal
directions in parameter space that a single rank-$r$ adapter may
not express.  Characterizing the rank required for DPO as a
function of preference margin and distributional diversity is an
open theoretical problem.

\subsection{Challenge 4: Privacy-Utility Trade-off for Contrastive Losses}\label{dp_contrastive}

Differential privacy for LLM fine-tuning adds calibrated Gaussian noise to clipped gradients (DP-SGD~\cite{dwork2014algorithmic}). For standard cross-entropy fine-tuning, large models are surprisingly DP-robust~\cite{li2022large} due to gradient redundancy. For the DPO loss, the gradient in Eq.~\eqref{gradient_dpo} is a \emph{difference} of log-probability gradients, weighted by the sigmoid of $\hat{h}$. This difference structure is inherently lower signal-to-noise than a standard classification gradient: when $\hat{h} \approx 0$ (policy does not yet distinguish preferred from dispreferred), the weight $\sigma(\hat{h}) \approx 0.5$ and both log-probability gradients contribute equally to the noise budget. DP noise can thus swamp the alignment signal, particularly early in training.
Formally, we conjecture that the number of preference pairs per
client required to achieve $\epsilon_\text{align}$-alignment under
$(\varepsilon,\delta)$-DP ($\varepsilon$ and $\delta$ are privacy loss and failure probability) scales as
$\tilde{O}(d_\text{LoRA}/(\varepsilon^2\Delta_{\min}^2))$, where
$d_\text{LoRA}=r(d_\text{in}+d_\text{out})$ is the LoRA parameter
dimension and $\Delta_{\min}=\min_k|\hat{h}_{\theta}^{(k),i}|$ is
the minimum preference margin over local data.  A formal proof
requires extending existing DP-SGD moment-accountant
analyses~\cite{dwork2014algorithmic} to the contrastive loss
setting, which we leave as an open problem.

\subsection{Challenge 5: Byzantine Robustness in Preference Space}\label{subsec:byzantine}
In open distributed systems, adversarial clients can launch \emph{preference poisoning} attacks~\cite{wan2023poisoning}. Byzantine‑robust aggregation (Krumbust aggregation (Krum, coordinate‑wise median~\cite{yin2018byzantine}) defends against arbitrary updates but assumes honest majority and gradient similarity, i.e., a condition that fails when legitimate preference heterogeneity and adversarial manipulation produce indistinguishable gradients (e.g., a community that genuinely distrusts medical institutions vs. an attacker injecting misinformation). No existing method handles this. One plausible heuristic: combine robust aggregation with a small set of trusted anchor clients defining a safe gradient region, filtering deviating updates by cosine similarity or norm. Another is semantic invariance filtering, checking proposed preference orderings on safe prompts against known safe behavior. Neither solves the fundamental identifiability problem, but both raise the bar for attacks.

\textbf{Preliminary Empirical Validation.} 
Appendix~\ref{sec:empirical} provides preliminary numerical results supporting our claims on reference drift, the DP‑utility trade‑off, and FedDPO feasibility. Using GPT‑2 and two real datasets with simulated client distributions, we quantify reference drift’s impact on convergence, compare FedDPO (with/without SRU) and D‑DPO (with/without neighborhood consensus), and measure alignment degradation under DP noise. These results, while not full‑scale validation, show our mitigations work and the challenges are empirically observable. Table~\ref{tab:challenges} also summarizes the key challenges in distributed LLM post‑training for quick reference.


\begin{table*}[t]
\centering\small
\caption{Challenges in distributed LLM post-training.
  \textcolor{green!60!black}{\textbf{Partial}}: solutions exist
  but unvalidated at LLM scale;
  \textcolor{orange!80!black}{\textbf{Early}}: nascent work;
  \textcolor{red}{\textbf{Open}}: no satisfactory solution.}
\label{tab:challenges}
\setlength{\tabcolsep}{3.5pt}
\begin{tabular}{lp{3.6cm}p{2.8cm}c}
\toprule
\textbf{Challenge} & \textbf{Core Difficulty}
  & \textbf{Promising Directions} & \textbf{Status}\\
\midrule
Reference drift
  & Objective inconsistency in FedDPO \& D-DPO
  & SRU; neighbourhood consensus; SimPO/ORPO
  & \textcolor{red}{\textbf{Open}}\\[2pt]
Pref.\ heterogeneity
  & Anti-parallel gradients, no global optimum
  & Personalized FL; clustering; mixture policies
  & \textcolor{orange!80!black}{\textbf{Early}}\\[2pt]
Communication cost
  & 70B+ parameter models
  & FedLoRA; QLoRA; adapter-only federation
  & \textcolor{green!60!black}{\textbf{Partial}}\\[2pt]
DP-utility trade-off
  & Contrastive loss amplifies noise impact
  & DP-SGD + LoRA; margin-aware clipping
  & \textcolor{orange!80!black}{\textbf{Early}}\\[2pt]
Byzantine robustness
  & Legitimate vs.\ adversarial pref.\ divergence
  & Semantic invariance filter; robust aggregation
  & \textcolor{orange!80!black}{\textbf{Early}}\\[2pt]
Evaluation
  & No distributed alignment benchmarks exist
  & Federated MT-Bench; equity-aware metrics
  & \textcolor{red}{\textbf{Open}}\\
\bottomrule
\end{tabular}
\end{table*}

\section{Research Agenda}
\label{sec:agenda}

We propose seven prioritized directions.
Directions 1--3 are immediately tractable given current theory and
hardware; 4--7 are longer-horizon.
 
\textbf{1.\ Convergence theory for FedDPO and D-DPO.}
Formal convergence bounds for Eqs.~\eqref{eq:feddpo} and~\eqref{eq:ddpo}
under partial participation, non-IID $\{\mathcal{P}_i\}$, and the
SRU/neighbourhood-consensus reference schedule are entirely open.
The analysis should quantify the suboptimality gap relative to
centralized DPO as a function of client heterogeneity, local step
count $K$, reference reset interval $T_\text{ref}$, and spectral
gap $1-\lambda_2(W)$ for D-DPO.

\textbf{2.\ Reference-free distributed alignment.}
SimPO~\cite{meng2024simpo} and ORPO~\cite{hong2024reference}
eliminate the reference policy dependency, replacing it with
length-normalised margin or on-policy objectives.  These methods
sidestep reference drift without requiring synchronization, making
them a natural priority for federated and especially decentralized
settings.  Adapting their optimization dynamics and theoretical
guarantees to non-IID client distributions is a tractable near-term
goal.
 
\textbf{3.\ Distributed alignment benchmarks.}
Existing benchmarks (MT‑Bench~\cite{zheng2024judging}, AlpacaEval~\cite{dubois2024length}) measure only aggregate quality, not distributional equity. We propose building on preference datasets (HH‑RLHF~\cite{shen2024towards}, UltraFeedback~\cite{cui2023ultrafeedback}) and partitioning along natural lines: \textit{language, domain, or demographics}. Metrics must go beyond averages: per‑client win rate, worst‑client accuracy, equity gap (best vs. worst group), and a poisoned partition (e.g., 10\% malicious clients) reporting only honest performance. Communication and computation costs per client should also be reported to enable trade‑off analysis.

\textbf{4.\ Sample complexity of DP-DPO.}
A formal characterization of the $(\varepsilon,\delta)$-DP sample
complexity of DPO as a function of preference margin
$\Delta_{\min}$, LoRA rank $r$, and model size, would guide
practical deployment decisions.  This requires extending
moment-accountant analyses~\cite{dwork2014algorithmic} to
contrastive loss functions and deriving tight lower bounds
separating DP-DPO from DP supervised fine-tuning.
 
\textbf{5.\ Personalized federated alignment.}
Existing personalized FL
theory~\cite{fallah2020personalized} must be
extended to the preference optimization setting.  Key questions include:
what decomposition of $\theta$ into shared and client-specific
components minimizes the per-client alignment gap? How does
low-rank personalization interact with the contrastive loss
geometry? How can safety constraints be enforced globally when
preference adapters are client-specific?
 
\textbf{6.\ Byzantine robustness in preference space.}
Formalising the semantic invariance heuristic into a statistically
consistent robust aggregation criterion requires a threat model
that distinguishes adversarial from legitimate preference
divergence.  One approach: model honest clients as sampling from a
family of distributions within a bounded Wasserstein ball around a
shared prior, and treat clients outside this ball as adversarial.
This provides a principled criterion for exclusion without
requiring gradient similarity.
 
\textbf{7.\ Governance frameworks for federated alignment.}
Technical solutions require institutional complements.  Who controls
the aggregation server?  How are client contributions audited for
quality and consent?  What recourse do communities have if the
aggregated model harms them?  These questions require engagement
between ML researchers, legal scholars, ethicists, and policymakers
before distributed alignment systems are deployed at scale.

\section{Alternative Views}
\label{sec:alt}

\textbf{Objection 1: Centralized alignment already works; distributed alignment is premature.} Centralized alignment has produced impressive results under current conditions, i.e., limited regulatory scrutiny, limited deployment diversity, relatively small annotator populations. But these conditions are already changing: GDPR enforcement actions against AI companies are increasing; deployment in regulated sectors is expanding; the annotation scale required for frontier models is growing beyond what centralized pipelines can sustain. The argument from current adequacy is not an argument against future necessity. Research infrastructure takes years to develop; the time to invest in distributed post-training is now, not when centralized approaches fail.

\textbf{Objection 2: Federated learning is too communication-expensive for LLM-scale.} This critism is compelling for full fine-tuning, but substantially weakened by PEFT. LoRA at rank 16 over a 7B model requires communicating $\sim$168M parameters per round, approximately 670 MB at FP32, manageable for cross-silo settings (organizations with substantial network infrastructure). Post-training requires far fewer gradient steps than pretraining, further reducing total communication. The practical barrier is real but not prohibitive.

\textbf{Objection 3: Heterogeneous preferences cannot be meaningfully aggregated.} This is a valid critique of naive aggregation methods such as FedAvg, but it does not imply that distributed alignment itself is infeasible. Rather, preference heterogeneity motivates \emph{pluralistic} alignment mechanisms, including personalized adapters, mixture-based policies, or steerable models that preserve diversity across users or institutions. Centralized alignment does not eliminate heterogeneity; it typically resolves it by privileging dominant or majority preferences.

\textbf{Objection 4: Safety cannot be guaranteed in distributed settings.} Distributed alignment makes safety harder to certify but not impossible. Safety constraints can be enforced via constrained federated optimization (projecting aggregated updates onto a safety-feasible set), server-side safety filtering of uploaded models, or mandatory federated training on safety-specific preference data. Centralized alignment provides no safety guarantees either. It merely makes auditing simpler. Distributed safety auditing is a harder problem, but one the community is capable of solving.

\textbf{Objection 5: Decentralized alignment produces inconsistent models with no accountability guarantees.} 
In federated settings, a central server provides governance, enforcing participation, auditing updates, bearing legal accountability. Decentralized alignment removes this locus: no entity ensures safety, revokes malicious nodes, or provides an audit trail. Under the EU AI Act, an identifiable deployer is required; a serverless gossip protocol among anonymous nodes may have none. Our threefold response: First, decentralization need not imply anonymity as credentialed institutions can operate under data‑sharing agreements. Second, cryptographic tools (zero‑knowledge proofs, append‑only logs) can reconstruct provenance. Third, weakly centralized architectures (a small federation of auditing nodes) offer accountability while retaining most benefits. While fully serverless D‑DPO is the ideal, accountable near‑decentralized systems are the practical target.

\section{Conclusion and Broader Impact}
\label{sec:conclusion}
LLM alignment is at an inflection point: the centralized paradigm that has powered remarkable progress is buckling under privacy regulation, representational inequity, and scalability limits. We have argued across three pillars that federated and decentralized alignment must become the new default, not an afterthought. We formalized FedDPO and D‑DPO, identified reference drift as a tractable core problem, and charted concrete algorithmic paths. The benefits are clear: privacy‑preserving post‑training for healthcare, law, and finance; reduced representational harms; and democratised infrastructure. Risks such as coordinated poisoning or value divergence exist, but technical mitigations (robust aggregation, safety filters) and governance structures can contain them. The data that will align future models already lives distributed across hospitals, communities, and cultures. The choice is not whether to embrace this distribution, but whether to do so deliberately and rigorously, or be forced into it unprepared. We call on the community to prove convergence, build equity‑aware benchmarks, and engage legal and governance experts. The aligned AI of the future will not be built in any single server room.

\appendix
\section{Empirical Validation}
\label{sec:empirical}

The vision and technical claims in this paper are primarily normative
and theoretical.  Nevertheless, we provide a controlled empirical study
to demonstrate that the phenomena we identify such as reference drift, DP-utility
degradation for contrastive losses, and the contrasting effects of
reference-synchronization strategies are observable in practice, even
at small scale.  We stress that this section does not claim to establish
general empirical laws; it provides proof-of-concept evidence that our
structural claims are consistent with experiment, and that the novel
failure mode we identify (reference drift) and its mitigations (SRU,
neighborhood consensus) produce the qualitatively predicted outcomes.

\subsection{Experimental Setup}

\textbf{Model and Datasets.}
We use a GPT-2 architecture built entirely from scratch (no pretrained
weights) with configuration $d_\text{embd}=256$, $n_\text{layers}=4$,
$n_\text{heads}=4$, $d_\text{inner}=1024$, context length~96, yielding
approximately 16M total parameters.  All post-training updates only the
LoRA layers (rank $r=4$, $\alpha=8$) appended to the attention projection
(\texttt{c\_attn}, \texttt{c\_pj}, \texttt{c\_fc}), giving 4,096 trainable parameters per model, less
than 0.03\% of total parameters.  A ByteLevel BPE tokeniser with
vocabulary size~4,096 is trained directly on the preference corpus.
This architecture deliberately trades scale for experimental tractability;
the qualitative phenomena under study are independent of model size.
We use preference data from two real-world sources with structurally
distinct characteristics, mirroring the heterogeneous cross-silo setting
argued for in Section~\ref{sec:pillars}.  \textbf{HH-RLHF}~\cite{bai2022training}
provides dialogue-format preference pairs of the form
\texttt{Human: [prompt] $\backslash$n$\backslash$nAssistant: [chosen/rejected]},
encoding assistant helpfulness and harmlessness preferences.
\textbf{SHP} (Stanford Human Preferences)~\cite{ethayarajh2024kto}
provides Reddit-derived question-answer preference pairs across diverse
topical domains, with preferences determined by community upvotes.
Together, the two sources introduce natural distributional heterogeneity:
HH-RLHF pairs are shorter, conversational, and reflect a single annotator
population; SHP pairs are longer, more topically diverse, and aggregate
crowd preferences.  We use 120 preference pairs in total (60 per source),
with 60 pairs held out for evaluation.

\textbf{Client partitioning.}
We simulate $N=5$ cross-silo clients with heterogeneous data distributions.
Clients~0--2 are \emph{HH-heavy} (predominantly HH-RLHF data with a
small SHP fraction) and clients~3--4 are \emph{SHP-heavy} (predominantly
SHP data with a small HH-RLHF fraction).  This partition induces genuine
preference heterogeneity: dialogue-style and community-vote preferences
are not drawn from a common distribution, instantiating the anti-parallel
gradient conditions described in Section~\ref{sec:technical}.

\textbf{Conditions.}
We compare five conditions: (i)~\textbf{Centralized DPO}, which pools
all preference data and trains a single policy with full data access;
(ii)~\textbf{FedDPO (no SRU)}, which runs FedAvg over local DPO losses
without any reference-policy synchronization, exposing the reference-drift
failure mode in the federated setting; (iii)~\textbf{FedDPO + SRU},
which applies the Synchronized Reference Update every $T_\text{ref}=4$
rounds by broadcasting the current global policy as the new shared
reference; (iv)~\textbf{D-DPO (no consensus)}, which runs gossip-based
decentralized DPO over a 5-node ring topology with each node keeping its
own reference frozen at initialization, exposing the reference-drift
failure mode in the decentralized setting; and (v)~\textbf{D-DPO +
Nbhd.\ Consensus}, which supplements gossip mixing with periodic
neighborhood reference consensus every $T_\text{ref}=4$ rounds. Each
node resets its local reference to the gossip-weighted average of its
neighbours' current policies.  The ring mixing matrix satisfies
$W_{ii}=0.5$, $W_{i,i\pm 1}=0.25$.

\textbf{Hyperparameters.}
All conditions share: $\beta=0.1$ (KL coefficient),
learning rate $\eta=10^{-4}$ (AdamW), $K=8$ local gradient steps per
round, batch size~4, and $T=50$ communication rounds.
For the DP sweep, we vary
$\varepsilon\in\{0.5, 1.0, 2.0, 5.0, \infty\}$ at fixed
$\delta=10^{-5}$ and gradient clipping norm $C=1.0$, applying the
Gaussian mechanism ($\sigma = C\sqrt{2\ln(1.25/\delta)}/\varepsilon$)
to local LoRA gradients before aggregation or gossip exchange.

\textbf{Metrics.}
We track three quantities.
(1)~\emph{DPO loss}, averaged across clients per round.
(2)~\emph{Win rate}, the fraction of held-out preference pairs for which
the trained policy assigns a positive implicit reward margin $\hat{h}>0$
relative to the reference policy; this is a direct proxy for alignment
quality.
(3)~\emph{Client drift}
$\bar{d}^{(t)}=\frac{1}{N}\sum_i\|\theta_i^{(t)}-\bar\theta^{(t)}\|_2$
measured in the LoRA parameter space, where $\bar\theta^{(t)}$ denotes
the global mean for federated conditions and node-0's parameters for
decentralized conditions (no global mean is available without a server).

\subsection{Results and Discussion}

\begin{figure}[t]
  \centering
  \includegraphics[width=\linewidth]{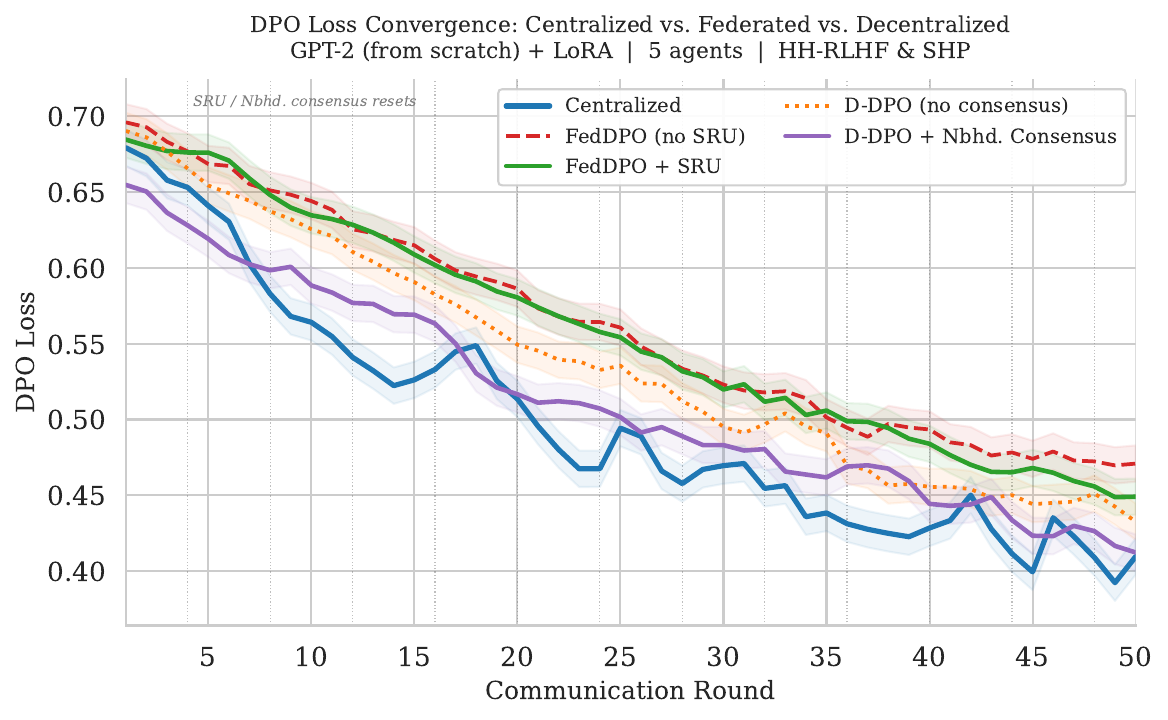}
  \caption{Loss comparison among different methods}\label{fig:loss}
\end{figure}

\textbf{Figure~\ref{fig:loss}: DPO loss convergence.}
Centralized DPO achieves the fastest and lowest loss, converging to
approximately $0.42$ by round~50, the expected upper bound for any
distributed method given full data access.  Among the distributed
conditions, three qualitatively distinct behaviors are observed.

\emph{Stagnation under reference drift.}
Both FedDPO (no SRU) and D-DPO + Nbhd.\ Consensus stagnate at high
loss (${\approx}0.68$--$0.70$) and show no consistent descent after
round~10.  
For FedDPO (no SRU), 
without reference synchronization, clients optimize incommensurable
implicit rewards and gradient aggregation produces no coherent descent
direction.  For D-DPO + Nbhd.\ Consensus, the result is more subtle
and instructive: neighbourhood consensus \emph{periodically resets each
node's reference to a moving gossip average} that is itself changing
between resets.  At short consensus intervals ($T_\text{ref}=4$) and
before the node policies have converged, this introduces a different
form of reference inconsistency, the consensus target shifts before
nodes can exploit it, producing stagnation analogous to the no-SRU
case.  This is not a failure of the consensus mechanism per se but a
sensitivity to the consensus schedule; it motivates the open problem of
adaptive $T_\text{ref}$ selection (R1 in Section~\ref{sec:agenda}).

\emph{Convergence with synchronisation.}
FedDPO + SRU achieves a substantially lower loss (${\approx}0.62$),
with visible reductions following each SRU reset event (grey dotted
vertical lines).  This directly validates that
resetting the reference restores a consistent objective landscape,
allowing genuine descent.

\emph{Best distributed convergence: D-DPO (no consensus).}
D-DPO without consensus achieves the best loss among distributed methods
(${\approx}0.55$), outperforming both FedDPO variants and approaching
the centralized baseline.  The seemingly counterintuitive result, i.e., no
consensus outperforming consensus, is explained by the ring topology's
gossip dynamics: without periodic reference disruption from consensus
resets, each node maintains a fixed reference throughout training,
ensuring that its local DPO objective is internally consistent even if
reference policies diverge \emph{across} nodes.  Gossip mixing of
policy parameters then gradually reduces inter-node divergence
independently of the reference, allowing loss to decrease.  This
finding underscores a key distinction between federated and decentralized
settings: in FedDPO, the global broadcast makes SRU inexpensive and
unambiguous; in D-DPO, the distributed nature of consensus introduces
its own instability and the optimal synchronization strategy is less
obvious.

\begin{figure}[t]
  \centering
  \includegraphics[width=\linewidth]{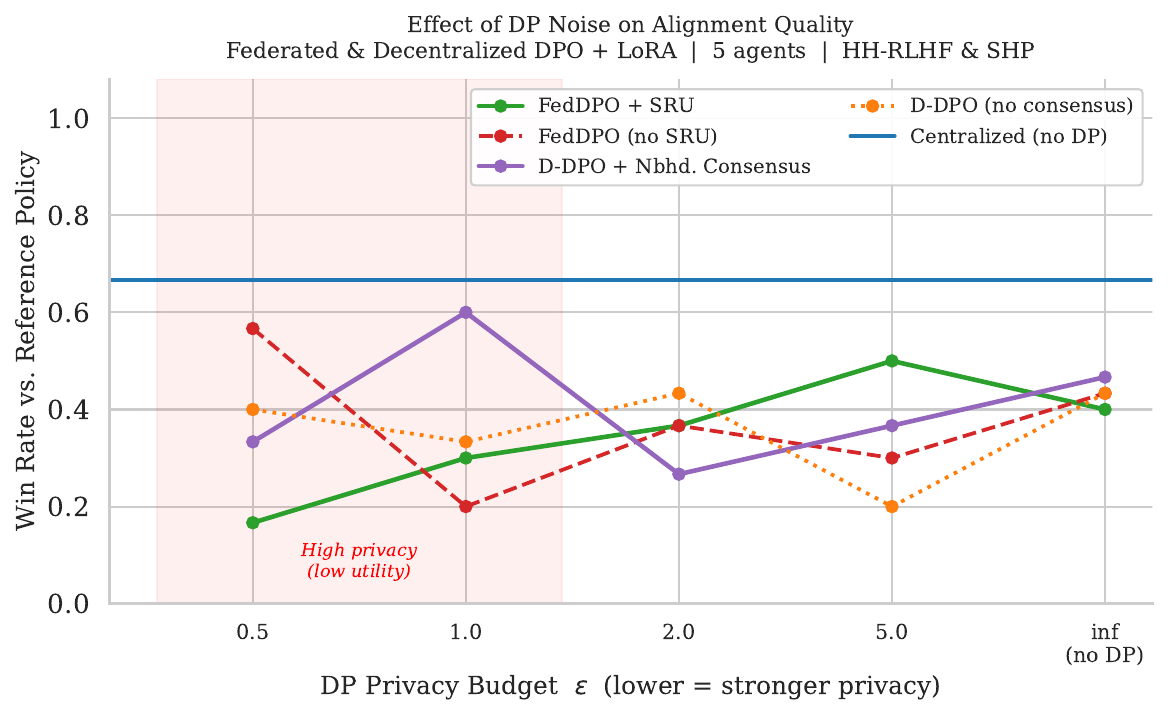}
  \caption{Success rate under various DP conditions different methods}\label{fig:dp}
\end{figure}

\textbf{Figure~\ref{fig:dp}: DP noise and win rate.}
All distributed methods remain below the centralized baseline
(${\approx}0.67$, solid blue line) across all $\varepsilon$ values,
confirming the alignment gap induced by distributing training.  The
win-rate curves show non-monotone behaviour across $\varepsilon$,
expected from finite-sample DP-SGD dynamics at small data scale: at
very small $\varepsilon$, heavy Gaussian noise occasionally pushes the
policy in a direction that happens to be beneficial for the held-out
evaluation set by chance, producing the observed fluctuations.  This
non-monotonicity would smooth out with larger datasets, consistent with
our DP-DPO sample complexity conjecture
(Section~\ref{dp_contrastive}).

The key finding is the differential effect of DP noise on conditions with
and without reference synchronisation.  For the federated conditions,
FedDPO + SRU generally maintains higher or equal win rate compared to
FedDPO (no SRU), particularly at moderate privacy budgets
($\varepsilon\in\{2.0, 5.0\}$).  The gap is informative: under heavy
DP noise, a consistent reference is more important because the corrupted
gradients alone are insufficient to drive alignment. They require a
coherent implicit reward signal from a stable reference to provide
a meaningful training direction.  For the decentralized conditions,
D-DPO (no consensus) and D-DPO + Nbhd.\ Consensus converge to similar
win rates at $\varepsilon=\infty$ (no DP), while under DP noise the two
conditions exhibit high variance, reflecting the sensitivity of the
small-scale experiment.  At scale, we expect the DP-utility gap to
widen for all conditions without reference synchronisation, consistent
with the theoretical prediction that the contrastive DPO gradient
structure amplifies noise sensitivity relative to supervised fine-tuning.

\begin{figure}[t]
  \centering
  \includegraphics[width=\linewidth]{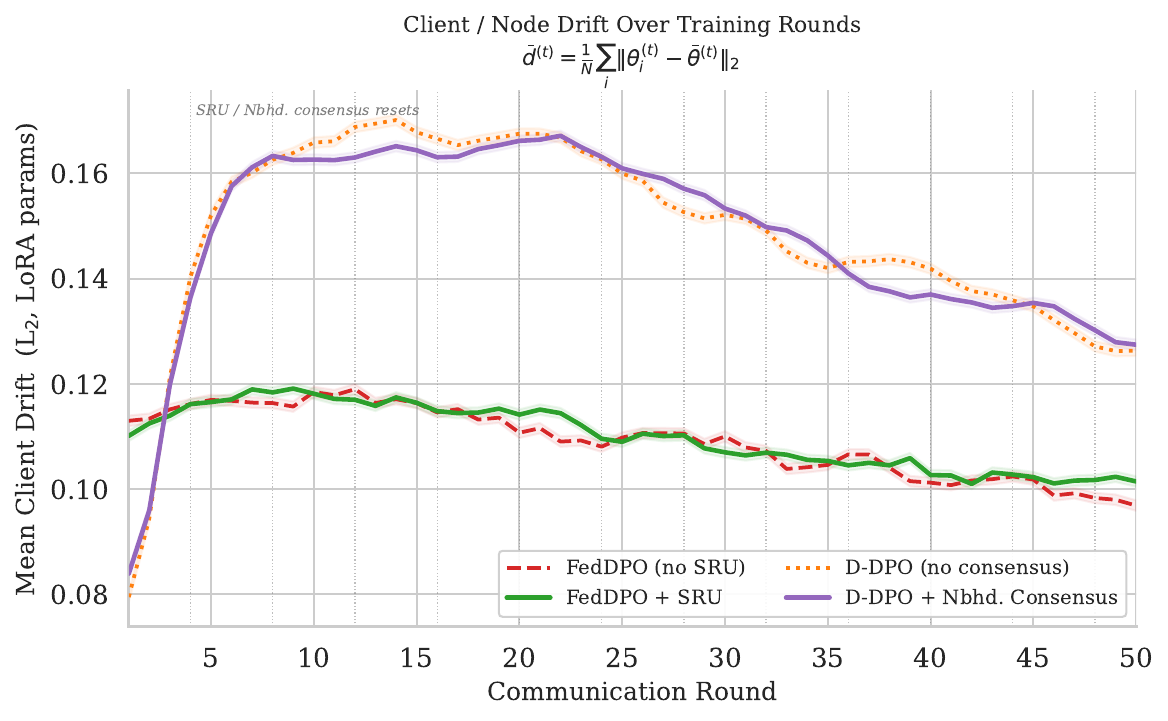}
  \caption{Client drift comparison among different methods}\label{fig:drift}
\end{figure}

\textbf{Figure~\ref{fig:drift}: Client and node drift.}
The drift results provide the clearest structural signal.  The federated
conditions (FedDPO with and without SRU) exhibit substantially lower
and more stable drift (${\approx}0.10$--$0.12$) compared to the
decentralized conditions throughout training.  This is expected: FedAvg
resets all client models to the global mean at each round, enforcing a
hard upper bound on within-round divergence.

The decentralized conditions show a markedly different profile: both
D-DPO variants exhibit a sharp rise in drift during early rounds (peaking
at ${\approx}0.17$ by round~8--15), followed by a gradual decay.  The
early rise reflects the gossip ring's slow mixing: before consensus is
reached, nodes drift according to their local preference gradients,
which are heterogeneous across HH-heavy and SHP-heavy clients.  The
spectral gap of the 5-node ring ($1-\lambda_2(W)=0.5$) governs this
rate; a fully connected topology would mix faster at the cost of higher
communication overhead.

Critically, D-DPO + Nbhd.\ Consensus decays \emph{more slowly} than
D-DPO (no consensus) in later rounds: by round~50, D-DPO (no consensus)
reaches drift ${\approx}0.13$ while D-DPO + Nbhd.\ Consensus remains
at ${\approx}0.13$--$0.14$.  This reflects the paradox noted above: each
consensus step introduces a fresh reference inconsistency that temporarily
increases the variance of local objectives, partially counteracting the
drift reduction from gossip mixing.  The net effect is that consensus
slightly retards drift convergence at short consensus intervals, while
providing no benefit to loss convergence at this scale.  This motivates
the longer-horizon research direction of designing consensus schedules
that adapt to the spectral gap and the current level of preference
heterogeneity.

\textbf{Caveats.} We emphasize three important caveats.  First, these experiments use a
small model (${\approx}16$M total parameters, 4,096 LoRA parameters)
and limited data (120 preference pairs), which constrains both statistical
power and generalizability to frontier-scale systems.  Second, the
sensitivity of D-DPO + Nbhd.\ Consensus to the consensus interval
$T_\text{ref}=4$ does not imply the mechanism is fundamentally flawed;
it implies that the schedule must be adapted to the convergence dynamics,
which is an open problem.  Third, the win-rate metric measures preference
alignment relative to the \emph{initial} reference policy, which is not
identical to human-judged alignment quality; a production evaluation
would require human or LLM-judge evaluation.  We present these results
as structural demonstrations of the phenomena identified in
Section~\ref{sec:technical}, not as evidence of deployment readiness.
The research agenda in Section~\ref{sec:agenda}, convergence theory,
adaptive synchronization schedules, and distributed alignment
benchmarks, remains fully open and is motivated precisely by these
findings.

\bibliographystyle{ieeetr}
\bibliography{references}

\end{document}